-----------------------------------------------------------------------------------------------------------------------

# A Comparative Study of AHP and Fuzzy AHP Method for Inconsistent Data

Md. Ashek-Al-Aziz[a]*, Sagar Mahmud[b], Md. Azizul Islam[c], Jubayer Al Mahmud[d], Khan Md. Hasib[e]

[a,b,c]*University of Development Alternative (UODA)*
[d]*University of Dhaka*
[e]*A hsanullah University of Science and Technology(AUST)*

**Abstract**

In various cases of decision analysis we use two popular methods – Analytical Hierarchical Process (AHP) and Fuzzy based AHP or Fuzzy AHP. Both the methods deal with stochastic data and can determine decision result through Multi Criteria Decision Making (MCDM) process. Obviously resulting values of the two methods are not same though same set of data is fed into them. In this research work, we have tried to observe similarities and dissimilarities between two methods' outputs. Almost same trend or fluctuations in outputs have been seen for both methods' for same set of input data which are not consistent. Both method outputs' ups and down fluctuations are same for fifty percent cases.

## 1. Introduction

Set Theory is a primary tool to describe classification of population data. Let's consider few classes of different age group people like Children Group for 0 to 18 years of age, Youth group for 18 to 25 years age, Middle age group for 20 to 55 years of age and old group for 50 to above age people. Mr. Zia is 56 years of old. For the proposition 'Mr. Zia is belonged to old group people' if he is 56 years of age then obviously the proposition is true. But for other group consideration the proposition will be false.

-----------------------------------------------------------------------

* Corresponding author.

This is the basic nature of crisp logic. Fuzzy logic makes the scenario different. If it is to be answered that how much old Mr. Zia is, then he is either not too old or very old. If Mr. Joy be 23 years of age then he is obviously young but how young he is. He is too young but if he would be 48 years of old then he would be partially young and merely old. This is vagueness and it is quantified considering a range 0 to 1 for assigning value to describe the vagueness. Let's say Mr. Zia's oldness is quantified by 0.4 whereas Mr Joy's youngness as 0.7 as 23 years of age. This is quite suitable technique for assessment in many cases. Let's consider another example. If we like to select a place for travelling among three places considering three criteria short distance (C1), safe travelling opportunity (C2) and expense (C3) then one of us may like to emphasize for minimum expense (C3) and least consider the safety of journey (C2) in short distance (C1) and select one place. But another one may have different view and want to put highest emphasis on distance (C1) and moderate priority on safety of journey (C2) and least consider the expense issue (C3). How this problem can be solved methodologically? Let's put the preferences for different criteria (C1 – C3) in percentage e.g. 0% for C1, 10% for C2 and 90% for C3 or 0.0 for C1, 0.1 for C2 and 0.9 for C3 respectively for prior case. Similarly 0.7, 0.2 and 0.1 for C1, C2 and C3 for later case. These data can be tabularize and suitable mathematical method can be applied so that common weight values can be obtained for these criteria which can be commonly used for selection of travelling place which should be acceptable for both. The method is further developed by determining the weight values of the criteria and taking decision using the weight values. We may like to assess the criteria with respect to one criterion i.e. regarding the criteria of short distance (C1) one should make assessment for other criteria like safe travelling (C1) and expense (C3) over C1 and the preferences can be expressed using numerical values in percentage or points in percentage. If this is done for all permutation of criteria then a pair wise preference comparison matrix can be formed and using a suitable mathematical method like AHP, the values of weights for such criteria can be determined. In AHP, such a pair wise comparison matrix values are summed up for each column, each elements are divided by each column sum then this first normalized matrix is averaged for each row to determine the weight values of criteria considered. We can easily understand that such methods deal with the stochastic data that holds some sort of uncertainty. Davoudi and Shykhvand (2012) states that due to the uncertainty in the judgments of participants, the crisp pair wise comparison in the conventional AHP is insufficient and imprecise to capture the right judgments. Therefore, a fuzzy logic is introduced in the pair-wise comparison of AHP in one of steps of the method [1]. Özdağoğl & özdağoğlu (2007) state that Fuzzy Analytical Hierarchy Process (FAHP) is a synthetic extension of classical AHP method when the fuzziness of the decision makers is considered [2, 3]. Kabir & Hasin (2011) demonstrated the advantages of capturing the vagueness of human thinking and to aid in solving the research problem through structured manner and simple process in a case of inventory classification. They have successfully identified the flexibility of incorporating more criteria or removing for specific implementation, different classification analyses can be done for different inventory records, application specific variable sets can be deployed and crisp comparison values can be substituted for fuzzy comparison values for optimization if fuzzy comparisons are not available in case [3]. Vayvay, Ozcan & Cruz-Cunha (2012) suggested from their experiments that both AHP and fuzzy AHP led to the same results, but neither of these considered the interactions within decision elements during the selection process [4]. But the major advantage of Fuzzy AHP methods stated by Mishra & Thakar (2012) that it can be used for both qualitative and quantitative criteria and also enables decision-makers to deal with inconsistent judgements systematically [5]. While incorporating decision engine into electronic procurement model, we could use normal

AHP method for selecting bidder, normalizing risk attributes' ratings and normalizing the maturity attributes' values but we have used fuzzy MCDM sometimes called fuzzy AHP instead of normal AHP. We have found that both the method uses stochastic process to deal with some uncertainty but major advantage of fuzzy MCDM over normal AHP is that maximizing the desired result or minimizing potential threats i.e. an optimization with Max-Min function is performed in fuzzy MCDM whereas normal AHP determines a deterministic result only. Here we have quoted form George J Klir & Bo Yuan [6]: "A decision is made under condition of risk, on the other hand, when the only available knowledge concerning the outcomes consists of their conditional probability distributions, one for each action. In this case, the decision making problem becomes an optimization problem of maximizing the expected utility. When probabilities of outcomes are not known, or may not even be relevant, and outcomes for each action are characterized only approximately, we say that decisions are made under uncertainty. This is the prime domain for fuzzy decision making." The motivation comes from this point that though we can implement fuzzy methods quite confidently but how much it holds similarities in behavior as seen in normal AHP method as both are decision tools. The aim of our research is to determine the output values of both AHP and Fuzzy AHP methods for the same inputs and observe the vibration of the successive results and compare them to have understanding their output fluctuations.

**Table 1:** Fuzzy pair wise decision criteria evaluation matrix for some criteria

|      | C 1  | C 2  | …   | C $n$ |
|------|------|------|-----|-------|
| C 1  | A11  | A1 2 | …   | A 1n  |
| C2   | A21  | A22  | …   | A2n   |
| .    | .    | .    |     | .     |
| .    | .    | .    |     | .     |
| .    | .    | .    |     | .     |
| Cn   | An1  | An2  | …   | Ann   |

$$A = \begin{pmatrix} a_{11} & a_{12} & \cdots & a_{1n} \\ a_{21} & a_{22} & \cdots & a_{2n} \\ \vdots & \vdots & \cdots & \vdots \\ a_{n1} & a_{n2} & & a_{nn} \end{pmatrix} \tag{1}$$

$$N = \begin{pmatrix} \dfrac{a_{11}}{\sum_{i=1}^{n} a_{i1}} & \dfrac{a_{12}}{\sum_{i=1}^{n} a_{i2}} & \cdots & \dfrac{a_{1n}}{\sum_{i=1}^{n} a_{in}} \\ \dfrac{a_{21}}{\sum_{i=1}^{n} a_{i1}} & \dfrac{a_{22}}{\sum_{i=1}^{n} a_{i2}} & \cdots & \dfrac{a_{2n}}{\sum_{i=1}^{n} a_{in}} \\ \vdots & \vdots & & \vdots \\ \dfrac{a_{n1}}{\sum_{i=1}^{n} a_{i1}} & \dfrac{a_{n2}}{\sum_{i=1}^{n} a_{i2}} & \cdots & \dfrac{a_{nn}}{\sum_{i=1}^{n} a_{in}} \end{pmatrix}$$

$$= \begin{pmatrix} N_{11} & N_{12} & \cdots & N_{1n} \\ N_{21} & N_{22} & \cdots & N_{2n} \\ \vdots & \vdots & \cdots & \vdots \\ N_{n1} & N_{n2} & & N_{nn} \end{pmatrix} \tag{2}$$

$$W = \begin{pmatrix} \frac{N_{11}+N_{12}\dots N_{1n}}{n} \\ \frac{N_{21}+N_{22}\dots N_{2n}}{n} \\ \vdots \\ \frac{N_{n1}+N_{n2}\dots N_{nn}}{n} \end{pmatrix} = \begin{pmatrix} w_1 \\ w_2 \\ \vdots \\ w_n \end{pmatrix} \qquad (3)$$

$$R = max\ (w_i) \qquad (4)$$

$$f\left(x_i, x_j\right) = \frac{f\left(x_i x_j\right)}{max\ [f\left(x_i, x_j\right), f\left(x_j, x_i\right)]} \qquad (5)$$

$$f^{/}(X_k) = min(X_i) \qquad (6)$$

$$N = \begin{pmatrix} \frac{a_{11}}{\sum_{i=1}^{n} a_{i1}} & \frac{a_{12}}{\sum_{i=1}^{n} a_{i2}} & \cdots & \frac{a_{1n}}{\sum_{i=1}^{n} a_{in}} \\ \frac{a_{21}}{\sum_{i=1}^{n} a_{i1}} & \frac{a_{22}}{\sum_{i=1}^{n} a_{i2}} & \cdots & \frac{a_{2n}}{\sum_{i=1}^{n} a_{in}} \\ \vdots & \vdots & & \vdots \\ \frac{a_{n1}}{\sum_{i=1}^{n} a_{i1}} & \frac{a_{n2}}{\sum_{i=1}^{n} a_{i2}} & \cdots & \frac{a_{nn}}{\sum_{i=1}^{n} a_{in}} \end{pmatrix}$$

$$= \begin{pmatrix} N_{11} & N_{12} & \cdots & N_{1n} \\ N_{21} & N_{22} & \cdots & N_{2n} \\ \vdots & \vdots & \cdots & \vdots \\ N_{n1} & N_{n2} & & N_{nn} \end{pmatrix} \qquad (2)$$

$$W = \begin{pmatrix} \frac{N_{11}+N_{12}\dots N_{1n}}{n} \\ \frac{N_{21}+N_{22}\dots N_{2n}}{n} \\ \vdots \\ \frac{N_{n1}+N_{n2}\dots N_{nn}}{n} \end{pmatrix} = \begin{pmatrix} w_1 \\ w_2 \\ \vdots \\ w_n \end{pmatrix}$$

(3)

$$R = max\ (w_i)$$

(4)

$$f\left(x_i, x_j\right) = \frac{f\left(x_i x_j\right)}{max\ [f\left(x_i, x_j\right), f\left(x_j, x_i\right)]}$$

(5)

$$f^{/}(X_k) = min(X_i)$$

(6)

$$R = max\ [f^{/}(X_k)]$$

(7)

## 2. AHP and Fuzzy AHP Method

AHP and Fuzzy AHP methods are useful tool for decision making when multiple criteria and multiple solution alternatives are present in a case. A pair wise comparison matrix is firstly formed if we have n numbers of criteria are available as in Table 1 and choice of preferences are substituted in the matrix. The Table 1 matrix is written as equation in eq (1). The matrix in normalized dividing each element of the matrix by column sum, then the new matrix N is formed and taking the row averages of N matrix, we get $w_i$ values of W row matrix where the final result is obtained by taking the highest values of $w_i$. This is all about AHP method discussed by Taha [7]. Both AHP and Fuzzy AHP deal with the stochastic data. Fuzzy logic allows quantification of vagueness and produce decision result. Fuzzy AHP is also called Fuzzy MPDM (Multi person decision making) or more specifically MPPC (Multi person preference criteria). After getting the fuzzy ratings, a pair wise comparison matrix like eq (1) is formed too and eq (5) does the first normalization. After first normalization, the minimum values of each row of first normalized matrix is extracted to form a new row matrix using eq (6) and finally highest value among the minimum values of each row as in eq (7) is the result determined by this method [6,8].

## 3. Data Analysis

A set of decision maker assessment using Fuzzy ratings are tabularized in Table 3 which is normalized using Fuzzy normalization in Table 4 and AHP normalization in Table 5. The minimum values of each row of Fuzzy normalized matrix in Table 4 are extracted and substituted in Table 6. Similarly the average values of each row of AHP normalized matrix from Table 5 are listed in Table 6 as well. These two result sets are compared and graphically showed in Figure 1. The whole process is repeated from Table 7 to Table 35 and Figure ure 2 to Figure 9 respectively. Various parameters are used in the Data Analysis because these data sets were used in procurement system analysis and the parameters remain unchanged. The explanations of these parameters are not essential in this research paper. We can see ups and downs of the curves in Figure ures and we have summarized the changes and similarity-dissimilarity in Table 36. In Table 37 and Table 38 we have tried to get concluding result of this research work. We should remember that all data sets taken into consideration for analysis is inconsistent i.e. all table data have consistency index (CI) values out of acceptable range. Various parameters are used from procurement cases here.

**Table 2:** Suggested numbers for maturity grading [4]

| $f(x_i, x_j)$ | Maturity weight $x_i$ of with respect to $x_j$ |
|---|---|
| 1 | Low maturity |
| 3 | Moderate maturity |
| 5 | High maturity |
| 7 | Very high maturity |
| 9 | Extra high maturity |
| 2,4,6,8 | Intermediate values between levels |

**Table 3:** Fuzzy rating of risk attributes by decision maker (nMax-61.72, CI=3.34, RI=1.72, CR=1.94)

| $f(x_i, x_j)$ | RELY | DURN | CPLX | CPIS | CADP | SCAP | WSZE | WSKL | SEXP | UMTG | SCED | PMEX | PDTH | RISK | RVOL |
|---|---|---|---|---|---|---|---|---|---|---|---|---|---|---|---|
| RELY | 1 | 3 | 7 | 9 | 7 | 9 | 3 | 5 | 5 | 5 | 3 | 1 | 3 | 9 | 5 |
| DURN | 5 | 1 | 3 | 5 | 5 | 3 | 7 | 5 | 3 | 5 | 9 | 3 | 3 | 3 | 9 |
| CPLX | 9 | 5 | 1 | 9 | 9 | 9 | 7 | 9 | 5 | 7 | 5 | 7 | 3 | 9 | 5 |
| CPIS | 3 | 3 | 5 | 1 | 1 | 3 | 3 | 5 | 3 | 1 | 3 | 5 | 3 | 7 | 1 |
| CADP | 9 | 1 | 1 | 1 | 1 | 1 | 3 | 3 | 5 | 5 | 3 | 3 | 1 | 7 | 1 |
| SCAP | 5 | 7 | 5 | 3 | 1 | 1 | 7 | 7 | 3 | 1 | 5 | 7 | 1 | 9 | 1 |
| WSZE | 1 | 9 | 7 | 1 | 1 | 3 | 1 | 9 | 3 | 3 | 5 | 1 | 1 | 3 | 1 |
| WSKL | 7 | 7 | 5 | 5 | 1 | 5 | 3 | 1 | 5 | 3 | 7 | 1 | 1 | 9 | 3 |
| SEXP | 1 | 1 | 3 | 1 | 1 | 5 | 3 | 7 | 1 | 1 | 1 | 3 | 1 | 3 | 1 |
| UMTG | 7 | 5 | 5 | 3 | 3 | 9 | 3 | 7 | 3 | 1 | 1 | 3 | 5 | 3 | 9 |
| SCED | 1 | 5 | 7 | 1 | 1 | 5 | 9 | 9 | 3 | 5 | 1 | 1 | 5 | 7 | 3 |
| PMEX | 3 | 1 | 1 | 3 | 1 | 7 | 3 | 5 | 7 | 5 | 3 | 1 | 7 | 1 | 1 |
| PDTH | 1 | 1 | 7 | 3 | 1 | 5 | 3 | 5 | 1 | 1 | 1 | 3 | 1 | 1 | 1 |
| RISK | 5 | 7 | 9 | 7 | 3 | 7 | 3 | 9 | 5 | 9 | 3 | 3 | 3 | 1 | 5 |
| RVOL | 5 | 1 | 3 | 1 | 7 | 3 | 1 | 1 | 5 | 1 | 5 | 3 | 5 | 7 | 1 |

**Table 4:** Normalized matrix of fuzzy ratings of Table 3 of Fuzzy Normalization

| $f(x_i, x_j)$ | RELY | DURN | CPLX | CPIS | CADP | SCAP | WSZE | WSKL | SEXP | UMTG | SCED | PMEX | PDTH | RISK | RVOL |
|---|---|---|---|---|---|---|---|---|---|---|---|---|---|---|---|
| RELY | 1 | 0.6 | 0.78 | 1 | 0.78 | 1 | 1 | 0.71 | 1 | 0.71 | 1 | 0.33 | 1 | 1 | 1 |
| DURN | 1 | 1 | 0.6 | 1 | 1 | 0.43 | 0.78 | 0.71 | 1 | 1 | 0.6 | 1 | 1 | 0.43 | 1 |
| CPLX | 0.33 | 1 | 1 | 1 | 1 | 1 | 1 | 1 | 1 | 1 | 1 | 1 | 0.43 | 1 | 1 |
| CPIS | 1 | 0.6 | 0.56 | 1 | 1 | 1 | 1 | 1 | 1 | 0.33 | 1 | 1 | 1 | 1 | 1 |
| CADP | 1 | 0.2 | 0.11 | 1 | 1 | 1 | 1 | 1 | 1 | 1 | 1 | 1 | 1 | 1 | 0.14 |
| SCAP | 0.56 | 1 | 0.56 | 1 | 1 | 1 | 1 | 1 | 0.6 | 0.11 | 1 | 1 | 0.2 | 1 | 0.33 |
| WSZE | 0.33 | 1 | 1 | 0.33 | 0.33 | 0.43 | 1 | 1 | 1 | 1 | 0.11 | 0.33 | 0.33 | 1 | 1 |
| WSKL | 1 | 1 | 0.56 | 1 | 0.33 | 0.71 | 0.33 | 1 | 0.71 | 0.43 | 0.11 | 0.2 | 0.2 | 1 | 1 |
| SEXP | 0.2 | 0.33 | 0.6 | 0.33 | 0.2 | 1 | 1 | 1 | 1 | 0.33 | 1 | 0.43 | 1 | 0.6 | 0.2 |
| UMTG | 1 | 1 | 0.71 | 1 | 0.6 | 1 | 1 | 1 | 1 | 1 | 1 | 1 | 1 | 1 | 1 |
| SCED | 0.33 | 0.55 | 1 | 0.33 | 0.33 | 1 | 1 | 1 | 1 | 1 | 1 | 0.33 | 1 | 1 | 0.6 |
| PMEX | 1 | 0.33 | 0.14 | 0.6 | 0.33 | 1 | 1 | 1 | 1 | 1 | 1 | 1 | 1 | 0.33 | 0.33 |
| PDTH | 0.33 | 0.33 | 1 | 1 | 1 | 1 | 1 | 1 | 1 | 0.33 | 0.6 | 0.43 | 1 | 0.33 | 0.2 |
| RISK | 0.56 | 1 | 1 | 1 | 0.43 | 0.78 | 1 | 1 | 1 | 1 | 0.43 | 1 | 1 | 1 | 0.71 |
| RVOL | 0.2 | 0.11 | 0.6 | 1 | 1 | 1 | 1 | 1 | 1 | 0.14 | 1 | 1 | 1 | 1 | 1 |

**Table 5:** Normalized matrix of Table 3 of AHP Normalization

| | RELY | DURN | CPLX | CPIS | CADP | SCAP | WSZE | WSKL | SEXP | UMTG | SCED | PMEX | PDTH | RISK | RVOL |
|---|---|---|---|---|---|---|---|---|---|---|---|---|---|---|---|
| RELY | 0.02 | 0.05 | 0.10 | 0.17 | 0.16 | 0.12 | 0.05 | 0.06 | 0.09 | 0.09 | 0.05 | 0.02 | 0.07 | 0.11 | 0.11 |
| DURN | 0.08 | 0.02 | 0.04 | 0.09 | 0.12 | 0.04 | 0.02 | 0.16 | 0.05 | 0.09 | 0.06 | 0.16 | 0.07 | 0.04 | 0.2 |
| CPLX | 0.14 | 0.09 | 0.01 | 0.17 | 0.21 | 0.12 | 0.12 | 0.10 | 0.09 | 0.13 | 0.09 | 0.15 | 0.08 | 0.11 | 0.11 |
| CPIS | 0.05 | 0.05 | 0.07 | 0.02 | 0.02 | 0.04 | 0.05 | 0.06 | 0.05 | 0.02 | 0.05 | 0.11 | 0.07 | 0.08 | 0.02 |
| CADP | 0.14 | 0.02 | 0.01 | 0.02 | 0.02 | 0.01 | 0.05 | 0.03 | 0.09 | 0.09 | 0.05 | 0.06 | 0.02 | 0.08 | 0.02 |
| SCAP | 0.08 | 0.12 | 0.07 | 0.06 | 0.02 | 0.01 | 0.12 | 0.08 | 0.05 | 0.02 | 0.09 | 0.15 | 0.02 | 0.11 | 0.02 |
| WSZE | 0.02 | 0.16 | 0.10 | 0.02 | 0.02 | 0.04 | 0.02 | 0.10 | 0.05 | 0.06 | 0.09 | 0.02 | 0.02 | 0.04 | 0.02 |
| WSKL | 0.11 | 0.12 | 0.07 | 0.09 | 0.02 | 0.07 | 0.05 | 0.01 | 0.09 | 0.06 | 0.12 | 0.02 | 0.02 | 0.11 | 0.07 |
| SEXP | 0.02 | 0.02 | 0.04 | 0.02 | 0.02 | 0.07 | 0.05 | 0.08 | 0.02 | 0.02 | 0.02 | 0.06 | 0.02 | 0.04 | 0.02 |
| UMTG | 0.11 | 0.09 | 0.07 | 0.06 | 0.07 | 0.12 | 0.05 | 0.08 | 0.05 | 0.02 | 0.05 | 0.11 | 0.07 | 0.11 | 0.16 |
| SCED | 0.02 | 0.09 | 0.10 | 0.02 | 0.02 | 0.07 | 0.15 | 0.10 | 0.05 | 0.09 | 0.02 | 0.02 | 0.12 | 0.08 | 0.07 |
| PMEX | 0.05 | 0.02 | 0.01 | 0.06 | 0.02 | 0.09 | 0.05 | 0.05 | 0.12 | 0.09 | 0.05 | 0.02 | 0.17 | 0.01 | 0.02 |
| PDTH | 0.02 | 0.02 | 0.10 | 0.06 | 0.02 | 0.07 | 0.05 | 0.06 | 0.02 | 0.02 | 0.02 | 0.06 | 0.02 | 0.01 | 0.02 |
| RISK | 0.08 | 0.12 | 0.13 | 0.13 | 0.07 | 0.09 | 0.05 | 0.10 | 0.09 | 0.17 | 0.05 | 0.06 | 0.07 | 0.01 | 0.11 |
| RVOL | 0.08 | 0.02 | 0.04 | 0.02 | 0.16 | 0.04 | 0.02 | 0.01 | 0.09 | 0.02 | 0.09 | 0.06 | 0.12 | 0.08 | 0.22 |

**Table 6:** Risk attributes' values generated by Fuzzy MCDM and AHP method

|  | RELY | DURN | CPLX | CPIS | CADP | SCAP | WSZE | WSKL | SEXP | UMTG | SCED | PMEX | PDTH | RISK | RVOL |
|---|---|---|---|---|---|---|---|---|---|---|---|---|---|---|---|
| AHP data | 0.085 | 0.083 | 0.114 | 0.051 | 0.05 | 0.069 | 0.052 | 0.069 | 0.034 | 0.081 | 0.068 | 0.057 | 0.038 | 0.09 | 0.058 |
| Fuzzy data | 0.33 | 0.429 | 0.33 | 0.33 | 0.11 | 0.11 | 0.11 | 0.11 | 0.2 | 0.6 | 0.33 | 0.143 | 0.2 | 0.429 | 0.11 |

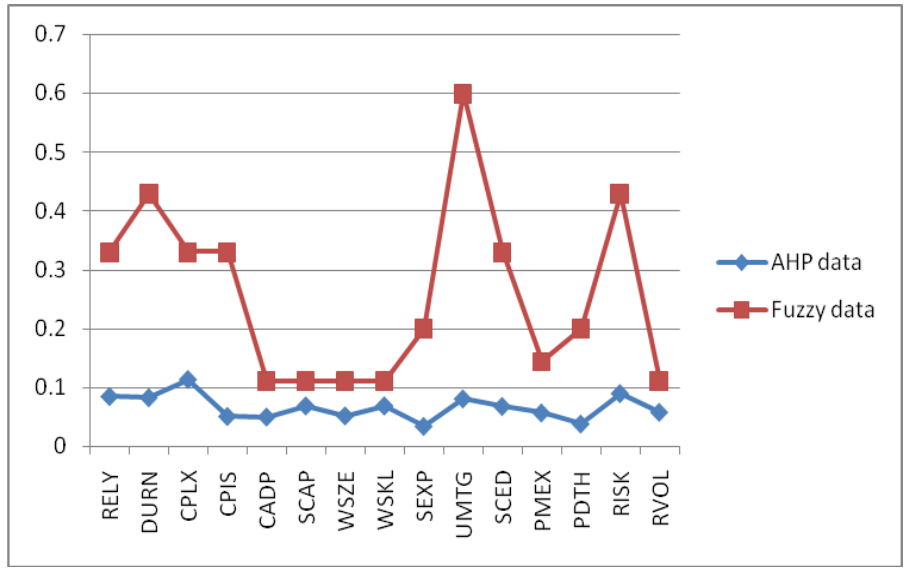

**Figure 1:** Comparison of AHP and fuzzy MCDM for risk attributes

**Table 7:** Fuzzy rating of procurement maturity identifier 'Customer' (nMax=18.86, CI=3.47, RI=1.19, CR=2.92)

|  | ENG | PIS | RMG | STF | SRT |
|---|---|---|---|---|---|
| ENG | 1 | 5 | 2 | 5 | 7 |
| PIS | 5 | 1 | 6 | 3 | 3 |
| RMG | 4 | 4 | 1 | 1 | 1 |
| STF | 6 | 8 | 1 | 1 | 8 |
| SRT | 3 | 1 | 8 | 8 | 1 |

**Table 8:** Normalized values of Table 7 matrix of Fuzzy Normalization

|  | ENG | PIS | RMG | STF | SRT |
|---|---|---|---|---|---|
| ENG | 1.0 | 1.0 | 0.5 | 0.83 | 1.0 |
| PIS | 1.0 | 1.0 | 0.38 | 1.0 |  |
| RMG | 1.0 | 0.67 | 1.0 | 1.0 | 0.12 |
| STF | 1.0 | 1.0 | 1.0 | 1.0 | 0.12 |
| SRT | 0.43 | 0.33 | 1.0 | 1.0 | 1.0 |

**Table 9:** Normalized values of Table 7 matrix of AHP Normalization

|       | ENG   | PIS   | RMG   | STF   | SRT   |
|-------|-------|-------|-------|-------|-------|
| ENG   | 0.053 | 0.263 | 0.111 | 0.278 | 0.350 |
| PIS   | 0.263 | 0.053 | 0.333 | 0.167 | 0.150 |
| RMG   | 0.211 | 0.211 | 0.056 | 0.056 | 0.050 |
| STF   | 0.316 | 0.421 | 0.056 | 0.056 | 0.400 |
| SRT   | 0.158 | 0.053 | 0.444 | 0.444 | 0.050 |

**Table 10:** Comparison data of AHP and Fuzzy MCDM against attributes of 'Customer'

|       | AHP data | Fuzzy data |
|-------|----------|------------|
| ENG   | 0.211    | 0.5        |
| PIS   | 0.193    | 0.38       |
| RMG   | 0.116    | 0.12       |
| STF   | 0.25     | 0.12       |
| SRF   | 0.23     | 0.33       |

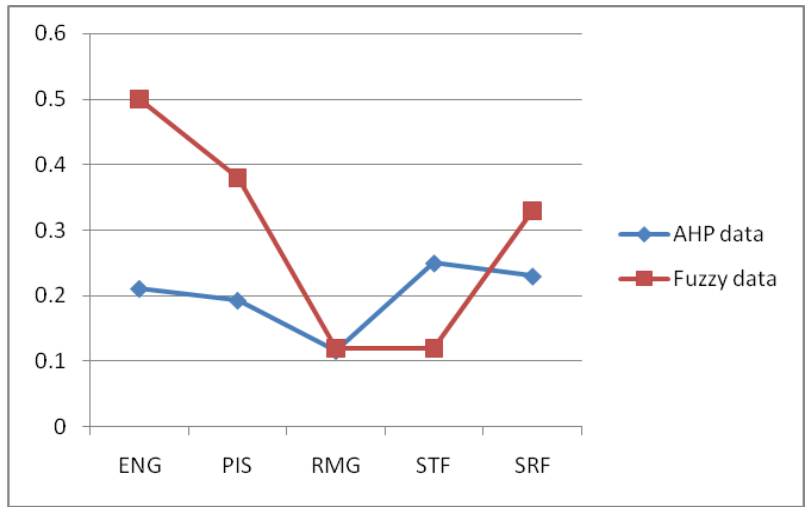

**Figure 2:** Comparison of AHP and fuzzy MCDM for PMM attribute 'Customer'

**Table 11:** Fuzzy rating of procurement maturity identifier 'Organization' (nMax=26.58, CI=3.26, RI=1.41, CR=2.31)

|       | BPC | BPN | EXS | MSN | SPN | STR | VSN |
|-------|-----|-----|-----|-----|-----|-----|-----|
| BPC   | 1   | 4   | 2   | 5   | 1   | 2   | 5   |
| BPN   | 5   | 1   | 6   | 7   | 3   | 5   | 8   |
| EXS   | 4   | 4   | 1   | 1   | 5   | 3   | 1   |
| MSN   | 6   | 8   | 1   | 1   | 5   | 1   | 3   |
| SPN   | 3   | 1   | 8   | 8   | 1   | 6   | 4   |
| STR   | 5   | 6   | 6   | 7   | 3   | 1   | 9   |
| VSN   | 4   | 4   | 3   | 1   | 5   | 4   | 1   |

**Table 12:** Normalized values of Table 8 matrix of Fuzzy Normalization

|      | BPC | BPN  | EXS | MSN  | SPN  | STR  | VSN  |
|------|-----|------|-----|------|------|------|------|
| BPC  | 1.0 | 0.8  | 0.5 | 0.83 | 0.33 | 0.4  | 1.0  |
| BPN  | 1.0 | 1.0  | 1.0 | 0.88 | 1.0  | 0.83 | 1.0  |
| EXS  | 1.0 | 0.67 | 1.0 | 1.0  | 0.62 | 0.5  | 0.33 |
| MSN  | 1.0 | 1.0  | 1.0 | 1.0  | 0.62 | 0.14 | 1.0  |
| SPN  | 1.0 | 0.33 | 1.0 | 1.0  | 1.0  | 1.0  | 0.8  |
| STR  | 1.0 | 1.0  | 1.0 | 1.0  | 0.5  | 1.0  | 1.0  |
| VSN  | 0.8 | 0.5  | 1.0 | 0.33 | 1.0  | 0.44 | 1.0  |

**Table 13:** Normalized values of Table 8 matrix of AHP Normalization

|      | BPC   | BPN   | EXS   | MSN   | SPN   | STR   | VSN   |
|------|-------|-------|-------|-------|-------|-------|-------|
| BPC  | 0.036 | 0.143 | 0.074 | 0.167 | 0.043 | 0.091 | 0.161 |
| BPN  | 0.179 | 0.036 | 0.222 | 0.233 | 0.130 | 0.227 | 0.258 |
| EXS  | 0.143 | 0.143 | 0.037 | 0.033 | 0.217 | 0.136 | 0.032 |
| MSN  | 0.214 | 0.286 | 0.037 | 0.033 | 0.217 | 0.045 | 0.097 |
| SPN  | 0.107 | 0.036 | 0.296 | 0.267 | 0.043 | 0.273 | 0.129 |
| STR  | 0.179 | 0.214 | 0.222 | 0.233 | 0.130 | 0.045 | 0.290 |
| VSN  | 0.143 | 0.143 | 0.111 | 0.033 | 0.217 | 0.182 | 0.032 |

**Table 14:** Comparison data of AHP and Fuzzy MCDM against maturity attributes of 'Organization'

|      | AHP   | Fuzzy |
|------|-------|-------|
|      | Data  | Data  |
| BPC  | 0.102 | 0.33  |
| BPN  | 0.184 | 0.83  |
| EXS  | 0.106 | 0.33  |
| MSN  | 0.133 | 0.14  |
| SPN  | 0.164 | 0.33  |
| STR  | 0.188 | 0.5   |
| VSN  | 0.123 | 0.33  |

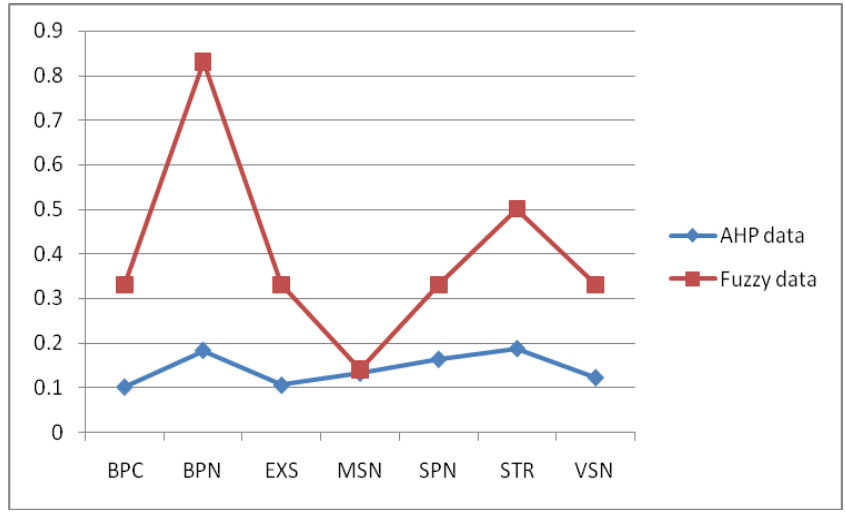

**Figure 3:** Comparison of AHP and fuzzy MCDM for PMM attribute 'Organization'

**Table 15:** Fuzzy rating of procurement maturity identifier 'Policy' (nMax=28.88, CI=3.65, RI=1.41, CR=2.58)

|     | AUL | BCP | DSP | PAT | PPY | PSD | RRN |
|-----|-----|-----|-----|-----|-----|-----|-----|
| AUL | 1   | 3   | 1   | 7   | 2   | 3   | 6   |
| BCP | 7   | 1   | 7   | 8   | 2   | 3   | 7   |
| DSP | 4   | 1   | 1   | 3   | 5   | 4   | 2   |
| PAT | 1   | 7   | 3   | 1   | 3   | 5   | 4   |
| PPY | 8   | 1   | 7   | 7   | 1   | 8   | 7   |
| PSD | 5   | 5   | 6   | 6   | 1   | 1   | 8   |
| RRN | 3   | 4   | 8   | 8   | 8   | 1   | 1   |

**Table 16:** Normalized values of Table 11 matrix of Fuzzy Normalization

|     | AUL | BCP  | DSP  | PAT | PPY  | PSD  | RRN  |
|-----|-----|------|------|-----|------|------|------|
| AUL | 1.0 | 0.43 | 0.25 | 1.0 | 0.25 | 0.6  | 1.0  |
| BCP | 1.0 | 1.0  | 1.0  | 1.0 | 1.0  | 0.6  | 1.0  |
| DSP | 1.0 | 0.14 | 1.0  | 1.0 | 0.71 | 0.67 | 0.25 |
| PAT | 1.0 | 0.88 | 1.0  | 1.0 | 0.43 | 0.83 | 0.5  |
| PPY | 1.0 | 0.5  | 1.0  | 1.0 | 1.0  | 1.0  | 0.88 |
| PSD | 1.0 | 1.0  | 1.0  | 1.0 | 0.12 | 1.0  | 1.0  |
| RRN | 0.5 | 0.57 | 1.0  | 1.0 | 1.0  | 0.12 | 1.0  |

**Table 17:** Normalized values of Table 11 of AHP Normalization

|     | AUL   | BCP   | DSP   | PAT   | PPY   | PSD  | RRN   |
|-----|-------|-------|-------|-------|-------|------|-------|
| AUL | 0.034 | 0.136 | 0.03  | 0.175 | 0.091 | 0.12 | 0.171 |
| BCP | 0.241 | 0.045 | 0.212 | 0.2   | 0.091 | 0.12 | 0.2   |
| DSP | 0.138 | 0.045 | 0.03  | 0.075 | 0.227 | 0.16 | 0.057 |
| PAT | 0.034 | 0.318 | 0.091 | 0.025 | 0.136 | 0.2  | 0.114 |
| PPY | 0.276 | 0.045 | 0.212 | 0.175 | 0.045 | 0.32 | 0.2   |
| PSD | 0.172 | 0.227 | 0.182 | 0.15  | 0.045 | 0.04 | 0.229 |
| RRN | 0.103 | 0.182 | 0.242 | 0.2   | 0.364 | 0.04 | 0.029 |

**Table 18:** Comparison data of AHP and Fuzzy MCDM against maturity attributes of 'Policy'

|  | AHP data | Fuzzy Data |
|---|---|---|
| AUL | 0.108 | 0.25 |
| BCP | 0.159 | 0.6 |
| DSP | 0.105 | 0.14 |
| PAT | 0.131 | 0.14 |
| PPY | 0.182 | 0.5 |
| PSD | 0.149 | 0.12 |
| RRN | 0.166 | 0.12 |

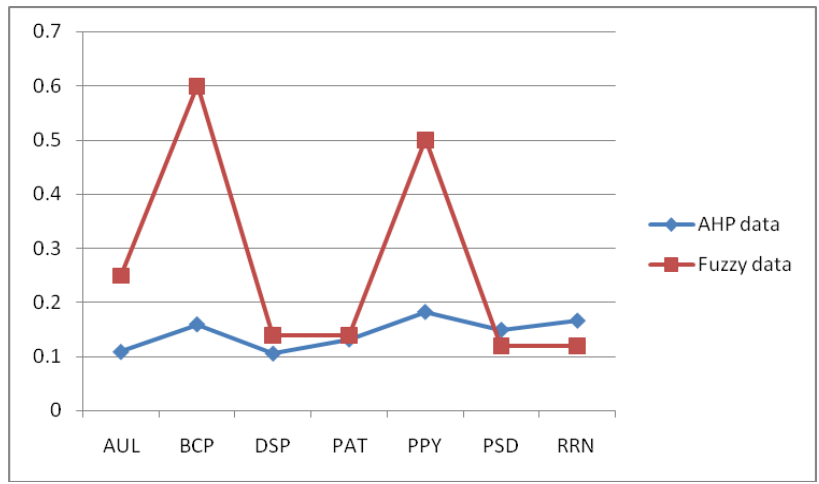

**Figure 4:** Comparison of AHP and fuzzy MCDM for PMM attribute 'Policy'

**Table 19:** Fuzzy rating of procurement maturity identifier 'Process' (nMax=28.57, CI=3.59, RI=1.41, CR=2.54)

|  | AUD | CBP | CRP | FRC | NGP | POG | SPL |
|---|---|---|---|---|---|---|---|
| AUD | 1 | 4 | 6 | 9 | 3 | 2 | 7 |
| CBP | 9 | 1 | 7 | 1 | 3 | 3 | 6 |
| CRP | 8 | 2 | 1 | 4 | 2 | 4 | 3 |
| FRC | 4 | 5 | 1 | 1 | 5 | 5 | 5 |
| NGP | 4 | 2 | 5 | 6 | 1 | 7 | 8 |
| POG | 1 | 6 | 7 | 6 | 9 | 1 | 1 |
| SPL | 7 | 3 | 7 | 5 | 1 | 2 | 1 |

**Table 20:** Normalized values of Table 14 matrix of Fuzzy Normalization

|     | AUD | CBP | CRP | FRC | NGP | POG | SPL |
|-----|-----|-----|-----|-----|-----|-----|-----|
| AUD | 1.0 | 0.44 | 0.75 | 1.0 | 0.75 | 1.0 | 1.0 |
| CBP | 1.0 | 1.0 | 1.0 | 0.2 | 1.0 | 0.5 | 1.0 |
| CRP | 1.0 | 0.29 | 1.0 | 1.0 | 0.4 | 0.57 | 0.43 |
| FRC | 0.44 | 1.0 | 0.25 | 1.0 | 0.83 | 0.83 | 1.0 |
| NGP | 1.0 | 0.67 | 1.0 | 1.0 | 1.0 | 0.78 | 1.0 |
| POG | 0.5 | 1.0 | 1.0 | 1.0 | 1.0 | 1.0 | 0.5 |
| SPL | 1.0 | 0.5 | 1.0 | 1.0 | 0.12 | 1.0 | 1.0 |

**Table 21:** Normalized values of Table 14 matrix of AHP Normalization

|     | AUD | CBP | CRP | FRC | NGP | POG | SPL |
|-----|-----|-----|-----|-----|-----|-----|-----|
| AUD | 0.029 | 0.174 | 0.176 | 0.281 | 0.125 | 0.083 | 0.226 |
| CBP | 0.265 | 0.043 | 0.206 | 0.031 | 0.125 | 0.125 | 0.194 |
| CRP | 0.235 | 0.087 | 0.029 | 0.125 | 0.083 | 0.167 | 0.097 |
| FRC | 0.118 | 0.217 | 0.029 | 0.031 | 0.208 | 0.208 | 0.161 |
| NGP | 0.118 | 0.087 | 0.147 | 0.188 | 0.042 | 0.292 | 0.258 |
| POG | 0.029 | 0.261 | 0.206 | 0.188 | 0.375 | 0.042 | 0.032 |
| SPL | 0.206 | 0.13 | 0.206 | 0.156 | 0.042 | 0.083 | 0.032 |

**Table 22:** Comparison data of AHP and fuzzy MCDM against maturity attributes of 'Process'

|     | AHP data | Fuzzy data |
|-----|----------|------------|
| AUD | 0.156 | 0.44 |
| CBP | 0.141 | 0.2 |
| CRP | 0.118 | 0.29 |
| FRC | 0.139 | 0.25 |
| NGP | 0.162 | 0.67 |
| POG | 0.162 | 0.5 |
| SPL | 0.122 | 0.12 |

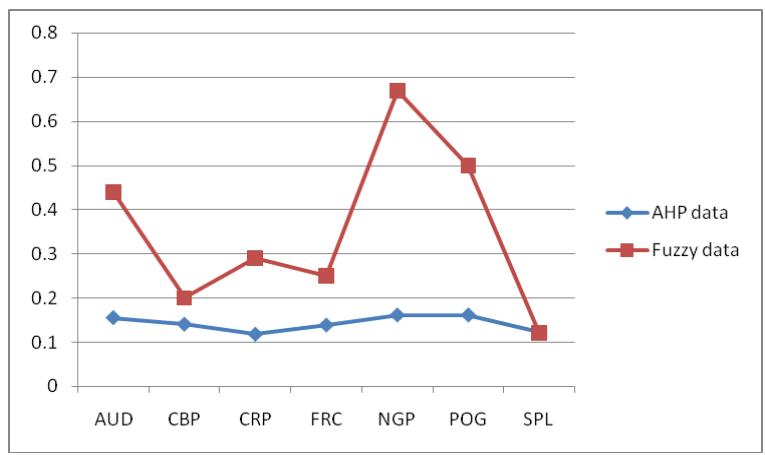

**Figure 5:** Comparison of AHP and fuzzy MCDM for PMM attribute 'Process'

**Table 23:** Fuzzy rating of procurement maturity identifier 'Staff' (nMax=38.62, CI=3.70, RI=1.54, CR=2.40)

|     | CRT | CTR | CEG | EEG | JQF | MPG | PFO | PRT | TRP |
|-----|-----|-----|-----|-----|-----|-----|-----|-----|-----|
| CRT | 1   | 3   | 7   | 5   | 1   | 2   | 6   | 3   | 7   |
| CTR | 4   | 1   | 5   | 6   | 8   | 7   | 8   | 5   | 1   |
| CEG | 8   | 2   | 1   | 4   | 2   | 4   | 3   | 1   | 9   |
| EEG | 4   | 5   | 1   | 1   | 5   | 5   | 5   | 4   | 5   |
| JQF | 5   | 5   | 3   | 1   | 1   | 8   | 4   | 7   | 6   |
| MGP | 1   | 6   | 7   | 6   | 9   | 1   | 1   | 7   | 2   |
| PFO | 9   | 9   | 7   | 1   | 3   | 3   | 1   | 6   | 2   |
| PRT | 1   | 4   | 6   | 9   | 3   | 2   | 7   | 1   | 3   |
| TRP | 4   | 3   | 6   | 7   | 8   | 3   | 6   | 4   | 1   |

**Table 24:** Normalized values of Table 17 matrix of Fuzzy Normalization

|     | CRT  | CTR  | CEG  | EEG  | JQF  | MPG  | PFO  | PRT  | TRP  |
|-----|------|------|------|------|------|------|------|------|------|
| CRT | 1.0  | 0.75 | 0.88 | 1.0  | 0.2  | 1.0  | 0.67 | 1.0  | 1.0  |
| CTR | 1.0  | 1.0  | 1.0  | 1.0  | 1.0  | 1.0  | 0.89 | 1.0  | 0.33 |
| CEG | 1.0  | 0.4  | 1.0  | 1.0  | 0.67 | 0.57 | 0.43 | 0.17 | 1.0  |
| EEG | 0.8  | 0.83 | 0.25 | 1.0  | 1.0  | 0.83 | 1.0  | 0.44 | 0.71 |
| JQF | 1.0  | 0.62 | 1.0  | 0.2  | 1.0  | 0.89 | 1.0  | 1.0  | 0.75 |
| MGP | 0.5  | 0.86 | 1.0  | 1.0  | 1.0  | 1.0  | 0.33 | 1.0  | 0.67 |
| PFO | 1.0  | 1.0  | 1.0  | 0.2  | 0.75 | 1.0  | 1.0  | 0.86 | 0.33 |
| PRT | 0.33 | 0.8  | 1.0  | 1.0  | 0.43 | 0.29 | 1.0  | 1.0  | 0.75 |
| TRP | 0.57 | 1.0  | 0.67 | 1.0  | 1.0  | 1.0  | 1.0  | 1.0  | 1.0  |

**Table 25:** Normalized values of Table 17 of AHP Normalization

|     | CRT   | CTR   | CEG   | EEG   | JQF   | MPG   | PFO   | PRT   | TRP   |
|-----|-------|-------|-------|-------|-------|-------|-------|-------|-------|
| CRT | 0.027 | 0.079 | 0.163 | 0.125 | 0.025 | 0.057 | 0.146 | 0.079 | 0.194 |
| CTR | 0.108 | 0.026 | 0.116 | 0.15  | 0.2   | 0.2   | 0.195 | 0.132 | 0.028 |
| CEG | 0.216 | 0.053 | 0.023 | 0.1   | 0.05  | 0.114 | 0.073 | 0.026 | 0.25  |
| EEG | 0.108 | 0.132 | 0.023 | 0.025 | 0.125 | 0.143 | 0.122 | 0.105 | 0.139 |
| JQF | 0.135 | 0.132 | 0.07  | 0.025 | 0.025 | 0.229 | 0.098 | 0.184 | 0.167 |
| MGP | 0.027 | 0.158 | 0.163 | 0.15  | 0.225 | 0.029 | 0.024 | 0.184 | 0.056 |
| PFO | 0.243 | 0.237 | 0.163 | 0.025 | 0.075 | 0.086 | 0.024 | 0.158 | 0.056 |
| PRT | 0.027 | 0.105 | 0.14  | 0.225 | 0.075 | 0.057 | 0.171 | 0.026 | 0.083 |
| TRP | 0.108 | 0.079 | 0.14  | 0.175 | 0.2   | 0.086 | 0.146 | 0.105 | 0.028 |

**Table 26:** Comparison data of AHP and Fuzzy MCDM against maturity attributes of 'Staff'

|     | AHP Data | Fuzzy data |
| --- | --- | --- |
| CRT | 0.1 | 0.2 |
| CTR | 0.128 | 0.33 |
| CEG | 0.101 | 0.17 |
| EEG | 0.102 | 0.25 |
| JQF | 0.118 | 0.2 |
| MGP | 0.113 | 0.33 |
| PFO | 0.118 | 0.2 |
| PRT | 0.101 | 0.29 |
| TRP | 0.119 | 0.57 |

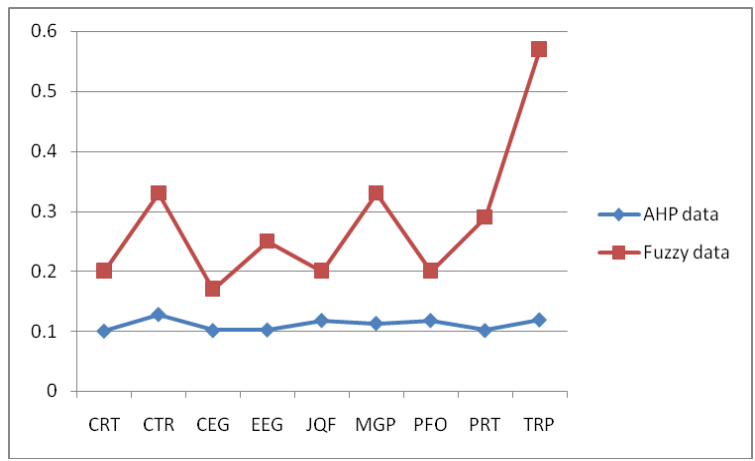

**Figure 6:** Comparison of AHP and fuzzy MCDM for PMM attribute 'Staff'

**Table 27:** Fuzzy rating of procurement maturity identifier 'Tools' (nMax=59.43, CI=3.49, RI=1.70, CR=2.06)

|     | CWA | CSS | CMS | CTL | RFX | EXW | PCD | PPO | RQS | RVA | RFT | TPR | VPS | VRM |
| --- | --- | --- | --- | --- | --- | --- | --- | --- | --- | --- | --- | --- | --- | --- |
| CWA | 1 | 7 | 8 | 4 | 7 | 3 | 7 | 1 | 6 | 7 | 7 | 3 | 7 | 5 |
| CSS | 6 | 1 | 9 | 1 | 4 | 2 | 5 | 9 | 9 | 7 | 4 | 2 | 5 | 6 |
| CMS | 1 | 3 | 1 | 6 | 8 | 2 | 5 | 1 | 4 | 6 | 8 | 2 | 5 | 4 |
| CTL | 9 | 3 | 2 | 1 | 4 | 5 | 1 | 4 | 3 | 6 | 4 | 5 | 1 | 2 |
| RFX | 7 | 3 | 7 | 5 | 1 | 2 | 9 | 9 | 1 | 2 | 5 | 5 | 3 | 1 |
| EXW | 4 | 2 | 5 | 6 | 8 | 1 | 3 | 3 | 6 | 5 | 1 | 6 | 7 | 6 |
| PCD | 8 | 2 | 5 | 4 | 2 | 4 | 1 | 2 | 7 | 2 | 2 | 5 | 4 | 2 |
| PPO | 4 | 5 | 1 | 2 | 5 | 5 | 3 | 1 | 5 | 7 | 5 | 1 | 2 | 5 |
| RQS | 5 | 5 | 3 | 1 | 7 | 8 | 9 | 3 | 1 | 7 | 4 | 5 | 3 | 2 |
| RVA | 4 | 5 | 1 | 2 | 5 | 5 | 5 | 4 | 5 | 1 | 4 | 3 | 1 | 1 |
| RFT | 5 | 5 | 3 | 1 | 7 | 8 | 4 | 7 | 6 | 6 | 1 | 1 | 9 | 8 |
| TPR | 8 | 2 | 5 | 4 | 2 | 4 | 3 | 3 | 1 | 2 | 5 | 1 | 5 | 7 |
| VPS | 2 | 5 | 5 | 5 | 4 | 6 | 9 | 2 | 3 | 1 | 7 | 8 | 1 | 9 |
| VRM | 3 | 1 | 8 | 8 | 3 | 5 | 1 | 3 | 3 | 6 | 8 | 2 | 5 | 1 |

**Table 28:** Normalized values of Table 20 matrix of Fuzzy Normalization

|     | CW A | CS S | CM S | CT L | RF X | EX W | PC D | PP O | RQ S | RV A | RF T | TP R | VP S | VR M |
| --- | --- | --- | --- | --- | --- | --- | --- | --- | --- | --- | --- | --- | --- | --- |
| CW A | 1.0 | 1.0 | 1.0 | 0.44 | 1.0 | 0.75 | 0.88 | 0.25 | 1.0 | 1.0 | 1.0 | 0.38 | 1.0 | 1.0 |
| CSS | 0.86 | 1.0 | 1.0 | 0.33 | 1.0 | 1.0 | 1.0 | 1.0 | 1.0 | 1.0 | 0.8 | 1.0 | 1.0 | 1.0 |
| CMS | 0.12 | 0.33 | 1.0 | 1.0 | 1.0 | 0.4 | 1.0 | 1.0 | 1.0 | 1.0 | 1.0 | 0.4 | 1.0 | 0.5 |
| CTL | 1.0 | 1.0 | 0.33 | 1.0 | 0.86 | 0.83 | 0.25 | 1.0 | 1.0 | 1.0 | 1.0 | 1.0 | 0.2 | 0.25 |
| RFX | 1.0 | 0.75 | 0.88 | 1.0 | 1.0 | 0.25 | 1.0 | 1.0 | 0.14 | 0.4 | 0.71 | 1.0 | 0.75 | 0.33 |
| EXW | 1.0 | 1.0 | 1.0 | 1.0 | 1.0 | 1.0 | 0.75 | 0.6 | 0.75 | 1.0 | 0.12 | 1.0 | 1.0 | 1.0 |
| PCD | 1.0 | 0.4 | 1.0 | 1.0 | 0.22 | 1.0 | 1.0 | 0.67 | 0.78 | 0.4 | 0.5 | 1.0 | 0.44 | 1.0 |
| PPO | 1.0 | 0.56 | 1.0 | 0.5 | 0.56 | 1.0 | 1.0 | 1.0 | 1.0 | 1.0 | 0.71 | 0.33 | 1.0 | 1.0 |
| RQS | 0.83 | 0.56 | 0.75 | 0.33 | 1.0 | 1.0 | 1.0 | 0.6 | 1.0 | 1.0 | 0.67 | 1.0 | 1.0 | 0.67 |
| RVA | 0.57 | 0.71 | 0.17 | 0.33 | 1.0 | 1.0 | 1.0 | 0.57 | 0.71 | 1.0 | 0.67 | 1.0 | 1.0 | 0.17 |
| RFT | 0.71 | 1.0 | 0.38 | 0.25 | 1.0 | 1.0 | 1.0 | 1.0 | 1.0 | 1.0 | 1.0 | 0.2 | 1.0 | 1.0 |
| TPR | 1.0 | 1.0 | 1.0 | 0.8 | 0.4 | 0.67 | 0.6 | 1.0 | 0.2 | 0.67 | 1.0 | 1.0 | 0.62 | 1.0 |
| VPS | 0.29 | 1.0 | 1.0 | 1.0 | 1.0 | 0.86 | 1.0 | 1.0 | 1.0 | 1.0 | 0.78 | 1.0 | 1.0 | 0.2 |
| VRM | 0.60 | 0.17 | 1.0 | 1.0 | 1.0 | 0.83 | 0.5 | 0.6 | 1.0 | 1.0 | 1.0 | 0.29 | 1.0 | 1.0 |

**Table 29:** Normalized values of Table 20 matrix of AHP Normalization

| | CWA | CSS | CMS | CTL | RFX | EXW | PCD | PPO | RQS | RVA | RFT | TPR | VPS | VRM |
|---|---|---|---|---|---|---|---|---|---|---|---|---|---|---|
| CWA | 0.015 | 0.143 | 0.127 | 0.08 | 0.104 | 0.05 | 0.108 | 0.019 | 0.1 | 0.108 | 0.108 | 0.061 | 0.121 | 0.085 |
| CSS | 0.09 | 0.02 | 0.143 | 0.02 | 0.06 | 0.033 | 0.077 | 0.173 | 0.15 | 0.108 | 0.062 | 0.041 | 0.086 | 0.102 |
| CMS | 0.015 | 0.061 | 0.016 | 0.12 | 0.119 | 0.033 | 0.077 | 0.019 | 0.067 | 0.092 | 0.123 | 0.041 | 0.086 | 0.068 |
| CTL | 0.134 | 0.061 | 0.032 | 0.02 | 0.06 | 0.083 | 0.015 | 0.077 | 0.05 | 0.092 | 0.062 | 0.102 | 0.017 | 0.034 |
| RFX | 0.104 | 0.061 | 0.111 | 0.1 | 0.015 | 0.038 | 0.133 | 0.173 | 0.017 | 0.032 | 0.077 | 0.102 | 0.052 | 0.017 |
| EXW | 0.06 | 0.041 | 0.079 | 0.1 | 0.119 | 0.017 | 0.046 | 0.058 | 0.1 | 0.075 | 0.015 | 0.122 | 0.121 | 0.102 |
| PCD | 0.119 | 0.041 | 0.079 | 0.08 | 0.03 | 0.067 | 0.015 | 0.038 | 0.117 | 0.031 | 0.031 | 0.102 | 0.069 | 0.034 |
| PPO | 0.06 | 0.102 | 0.016 | 0.04 | 0.075 | 0.073 | 0.046 | 0.019 | 0.083 | 0.108 | 0.077 | 0.02 | 0.034 | 0.085 |
| RQS | 0.075 | 0.102 | 0.042 | 0.02 | 0.104 | 0.133 | 0.138 | 0.058 | 0.017 | 0.102 | 0.062 | 0.102 | 0.052 | 0.034 |
| RVA | 0.06 | 0.102 | 0.016 | 0.04 | 0.075 | 0.083 | 0.077 | 0.077 | 0.083 | 0.015 | 0.062 | 0.061 | 0.017 | 0.017 |
| RFT | 0.075 | 0.102 | 0.048 | 0.02 | 0.104 | 0.133 | 0.062 | 0.135 | 0.1 | 0.092 | 0.015 | 0.02 | 0.155 | 0.136 |
| TPR | 0.119 | 0.041 | 0.079 | 0.08 | 0.03 | 0.067 | 0.046 | 0.058 | 0.017 | 0.031 | 0.077 | 0.02 | 0.086 | 0.119 |
| VPS | 0.03 | 0.102 | 0.079 | 0.1 | 0.06 | 0.1 | 0.138 | 0.038 | 0.05 | 0.015 | 0.108 | 0.163 | 0.017 | 0.153 |
| VRM | 0.045 | 0.02 | 0.127 | 0.16 | 0.045 | 0.083 | 0.015 | 0.058 | 0.05 | 0.092 | 0.123 | 0.041 | 0.086 | 0.017 |

**Table 30:** Comparison data of AHP and fuzzy MCDM against maturity attributes of 'Tools'

| | AHP data | Fuzzy data |
|---|---|---|
| CWA | 0.088 | 0.25 |
| CSS | 0.083 | 0.33 |
| CMS | 0.067 | 0.12 |
| CTL | 0.06 | 0.2 |
| RFX | 0.074 | 0.14 |
| EXW | 0.077 | 0.12 |
| PCD | 0.061 | 0.22 |
| PPO | 0.061 | 0.33 |
| RQS | 0.075 | 0.33 |
| RVA | 0.056 | 0.17 |
| RFT | 0.086 | 0.2 |
| TPR | 0.062 | 0.2 |
| VPS | 0.082 | 0.2 |
| VRM | 0.069 | 0.17 |

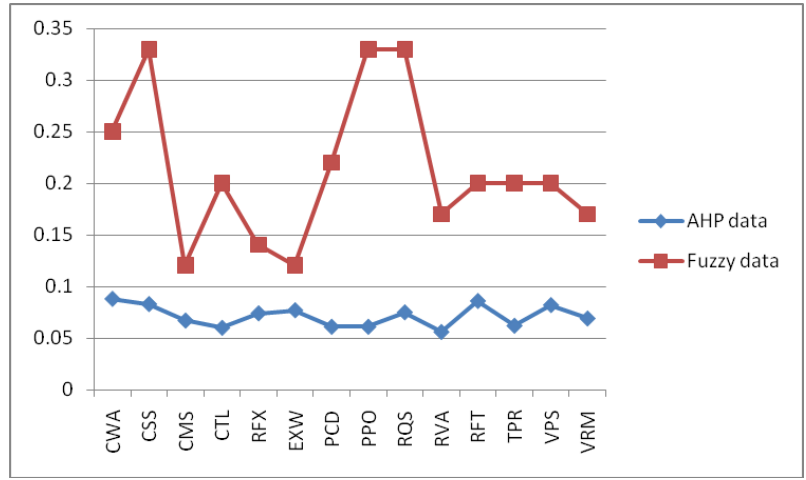

**Figure 7:** Comparison of AHP and fuzzy MCDM for PMM attribute 'Tools'

**Table 31:** Fuzzy rating of procurement maturity identifier 'Vendors' (nMax=25.03, CI=3.81, RI=1.32, CR=2.89)

|  | AVL | MMS | VCN | VQN | VRN | VRG |
|---|---|---|---|---|---|---|
| AVL | 1 | 5 | 4 | 2 | 4 | 3 |
| MMS | 5 | 1 | 2 | 5 | 5 | 5 |
| VCN | 5 | 3 | 1 | 7 | 8 | 4 |
| VQN | 6 | 7 | 6 | 1 | 9 | 1 |
| VRN | 9 | 7 | 1 | 3 | 1 | 6 |
| VRG | 4 | 6 | 9 | 3 | 2 | 1 |

**Table 32:** Normalized values of Table 26 matrix of Fuzzy Normalization

|  | AVL | MMS | VCN | VQN | VRN | VRG |
|---|---|---|---|---|---|---|
| AVL | 1.0 | 1.0 | 0.8 | 0.33 | 0.44 | 0.75 |
| MMS | 1.0 | 1.0 | 0.67 | 0.71 | 0.71 | 0.83 |
| VCN | 1.0 | 1.0 | 1.0 | 1.0 | 1.0 | 0.44 |
| VQN | 1.0 | 1.0 | 0.86 | 1.0 | 1.0 | 0.33 |
| VRN | 1.0 | 1.0 | 0.12 | 0.33 | 1.0 | 1.0 |
| VRG | 1.0 | 1.0 | 1.0 | 1.0 | 0.33 | 1.0 |

**Table 33:** Normalized values of Table 26 matrix of AHP Normalization

|  | AVL | MMS | VCN | VQN | VRN | VRG |
|---|---|---|---|---|---|---|
| AVL | 0.033 | 0.172 | 0.174 | 0.095 | 0.138 | 0.15 |
| MMS | 0.167 | 0.034 | 0.087 | 0.238 | 0.172 | 0.25 |
| VCN | 0.167 | 0.103 | 0.043 | 0.333 | 0.276 | 0.2 |
| VQN | 0.2 | 0.241 | 0.261 | 0.048 | 0.31 | 0.05 |
| VRN | 0.3 | 0.241 | 0.043 | 0.143 | 0.034 | 0.3 |
| VRG | 0.133 | 0.207 | 0.391 | 0.143 | 0.069 | 0.05 |

**Table 34:** Comparison data of AHP and fuzzy MCDM against maturity attributes of 'Vendors'

|     | AHP data | Fuzzy data |
| --- | --- | --- |
| AVL | 0.127 | 0.33 |
| MMS | 0.158 | 0.67 |
| VCN | 0.187 | 0.44 |
| VQN | 0.185 | 0.33 |
| VRN | 0.177 | 0.12 |
| VRG | 0.166 | 0.33 |

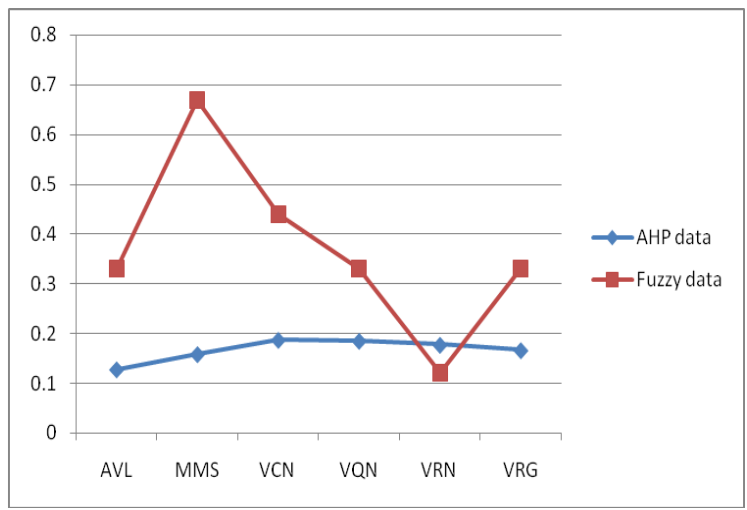

**Figure 8:** Comparison of AHP and fuzzy MCDM for PMM attribute 'Vendor'

**Table 35:** Comparison data of AHP and fuzzy MCDM of decision values of specific PMM attributes

|     | AHP decision values | Fuzzy decision values |
| --- | --- | --- |
| Customer | 0.25 | 0.5 |
| Organization | 0.188 | 0.83 |
| Policy | 0.182 | 0.6 |
| Process | 0.162 | 0.67 |
| Staff | 0.128 | 0.57 |
| Tools | 0.088 | 0.33 |
| Vendors | 0.187 | 0.67 |

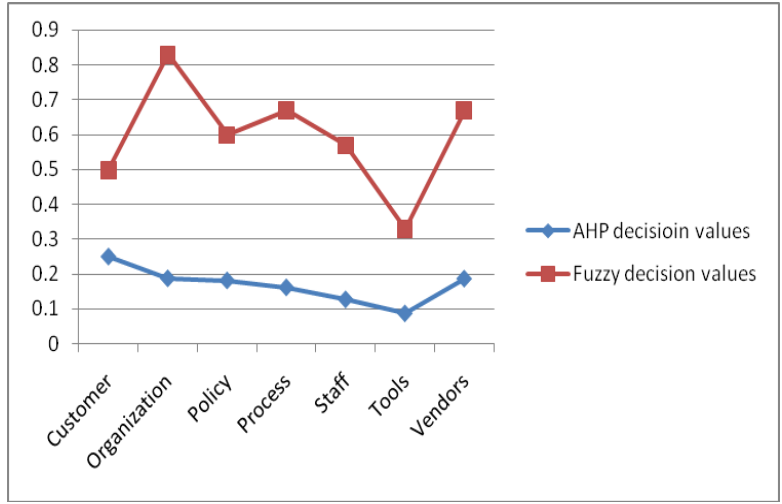

**Figure 9:** Comparison of AHP and fuzzy MCDM for PMM attributes

**Table 36:** Observation records of AHP and Fuzzy MCDM comparison graphs

|  | Increase in AHP, Increase in Fuzzy | Increase in AHP, Decrease in Fuzzy | Decrease in Fuzzy, Increase in AHP | Decrease in AHP, decrease in Fuzzy | Increase/Decrease in AHP, Fuzzy unchanged | Increase/Decrease in Fuzzy, AHP unchanged | Total observations |
|---|---|---|---|---|---|---|---|
| Risk Identifier | 2 | 2 | 3 | 3 | 4 | × | 14 |
| Customer | × | × | 1 | 2 | 1 | × | 4 |
| Organization | 3 | 1 | × | 2 | × | × | 6 |
| Policy | 2 | × | × | 2 | 2 | × | 6 |
| Process | 1 | 1 | 1 | 2 | × | 1 | 6 |
| Staff | 3 | 2 | 2 | 1 | × | × | 8 |
| Tools | 1 | 2 | 3 | 3 | 3 | 1 | 13 |
| Vendors | 1 | 1 | 1 | 2 | × |  | 5 |
| Few decision ratings | 1 | × | 2 | 3 | × | × | 6 |

**Table 37:** Observation summery of AHP and Fuzzy MCDM comparison curves

| | |
|---|---|
| Increase in AHP, increase in Fuzzy MCDM | 20.59% |
| Increase in AHP, decrease in Fuzzy MCDM | 13.24% |
| Decrease in AHP, increase in Fuzzy MCDM | 19.12% |
| Decrease in AHP, decrease in Fuzzy MCDM | 29.41% |
| Increase or decrease in AHP, Fuzzy remain unchanged | 14.71% |
| Increase or decrease in Fuzzy, AHP remain unchanged | 2.94% |

**Table 38:** Observation result of AHP and Fuzzy MCDM comparison

| | |
|---|---|
| Increase or decrease in AHP makes increase or decrease in Fuzzy respectively | 50% |
| Increase or decrease in AHP makes reverse swing in Fuzzy | 32.36% |
| Either AHP or Fuzzy remain unchanged for any slope of Fuzzy or AHP respectively | 17.64% |

## 4. Conclusion

A summary of total observations are snapped. We have noticed that fuzzy curve is quite similar in nature with AHP curve characteristics. When fuzzy data is increased, we see that AHP data is successively increased and decreased for fuzzy decrease and the vibration of both the curve is same for many samples for most of the cases except some few. The rise of Fuzzy data makes the rise in AHP and vice versa is secured for 50% of the cases we considered.

## Acknowledgements

I am greatfull to Dr. Abdullahil Azeem who advised me to work on the comparative study of AHP and Fuzzy AHP methods and produce a data analysis results. I like to thank to my friends who encouraged me to make Excel data sheet and charts to produce the data analysis more meaningful.